\documentclass[3p, twocolumn]{elsarticle}

\usepackage{amssymb}
\usepackage{tabularx}
\usepackage{booktabs}
\usepackage{makecell}
\usepackage{subcaption}
\usepackage{adjustbox}

\usepackage{ulem}
\usepackage{array}
\usepackage{graphicx}

\usepackage[binary-units = true]{siunitx}
\DeclareSIUnit\px{px}
\global\nopreprintlinetrue

\newcolumntype{L}[1]{>{\raggedright\arraybackslash}p{#1}}

\begin{document}

\begin{frontmatter}

\title{Classifying Whole Slide Images: What Matters?}


\author[inst1]{Long Nguyen}
\author[inst1]{Aiden Nibali}
\author[inst1]{Joshua Millward}
\author[inst1]{Zhen He\corref{mycorrespondingauthor}}
\cortext[mycorrespondingauthor]{Corresponding author}

\affiliation[inst1]{organization={Department of Computer Science and Information Technology},
            addressline={La Trobe University}, 
            city={Bundoora},
            postcode={3086}, 
            state={VIC},
            country={Australia}}

\begin{abstract}

Recently there have been many algorithms proposed for the classification of very high resolution whole slide images (WSIs). These new algorithms are mostly focused on finding novel ways to combine the information from small local patches extracted from the slide, with an emphasis on effectively aggregating more global information for the final predictor. In this paper we thoroughly explore different key design choices for WSI classification algorithms to investigate what matters most for achieving high accuracy. Surprisingly, we found that capturing global context information does not necessarily mean better performance. A model that captures the most global information consistently performs worse than a model that captures less global information. In addition, a very simple multi-instance learning method that captures no global information performs almost as well as models that capture a lot of global information. These results suggest that the most important features for effective WSI classification are captured at the local small patch level, where cell and tissue micro-environment detail is most pronounced. Another surprising finding was that unsupervised pre-training on a larger set of 33 cancers gives significantly worse performance compared to pre-training on a smaller dataset of 7 cancers (including the target cancer). We posit that pre-training on a smaller, more focused dataset allows the feature extractor to make better use of the limited feature space to better discriminate between subtle differences in the input patch.

\end{abstract}




\end{frontmatter}


\section{Introduction}
\label{sec:intro}
The application of computer vision techniques to digital pathology has the potential to become a transformative force in the field of medical diagnostics, with experts agreeing that the routine use of AI tools in future pathology labs is almost assured~\cite{berbis2023computational}. By automating the analysis of whole slide images (WSIs) it will be possible to enhance the work of clinical histopathologists, ultimately leading to more efficient personalised treatment planning for patients. 

Many recent works \cite{hipt,adnan2020graph,guan2022node,stegmuller2023scorenet,li2021dual,shao2021transmil,vu2023handcrafted} focus on applying deep learning approaches to solve the weakly supervised whole slide image classification problem. The WSI is taken as input, and the model is trained to output a single label such as the cancer sub-type, metastasised versus normal lymph node, presence of certain genes, etc. 

A major practical challenge when working with WSIs is their extremely high dimensionality, with a single image reaching spatial extents in the order of $\SI{100000}{\px} \times \SI{100000}{\px}$. It is impossible to directly feed all pixels from such images into a neural network at once. The majority of recent methods \cite{hipt,adnan2020graph} first break the image into small tiles (e.g., $\SI{256}{\px} \times \SI{256}{\px}$ patches) and then represent each tile using a small 1D embedding (e.g. a 384-dimensional vector). These feature vectors are typically generated using models pre-trained on natural images from ImageNet or a large collection of pan-cancer WSIs. Typically, the ImageNet pre-trained model weights are computed via the supervised task of image classification. In contrast, models pre-trained on large WSI collections are usually trained without annotations by using self-supervised learning techniques. Either way, the tile-based feature vectors are combined using various methods to arrive at a single whole slide prediction.  

Although the tile-based approach already allows later stages of the model to operate at a higher level by working on $\SI{256}{\px} \times \SI{256}{\px}$ patches instead of individual pixels at a time, most recent work advocates the importance of incorporating even more global structural information when classifying WSIs. To capture global structure information, previous works have connected patches in a graph\cite{adnan2020graph,guan2022node}, used a mixture of high resolution and low resolution images\cite{stegmuller2023scorenet,li2021dual}, applied self attention with position encoding on image patches\cite{shao2021transmil,vu2023handcrafted}, and built a hierarchical representation of WSIs using separate vision transformers at different levels of the  hierarchy\cite{hipt}.  

\begin{figure*}
    \centering
    \includegraphics[width=\textwidth]{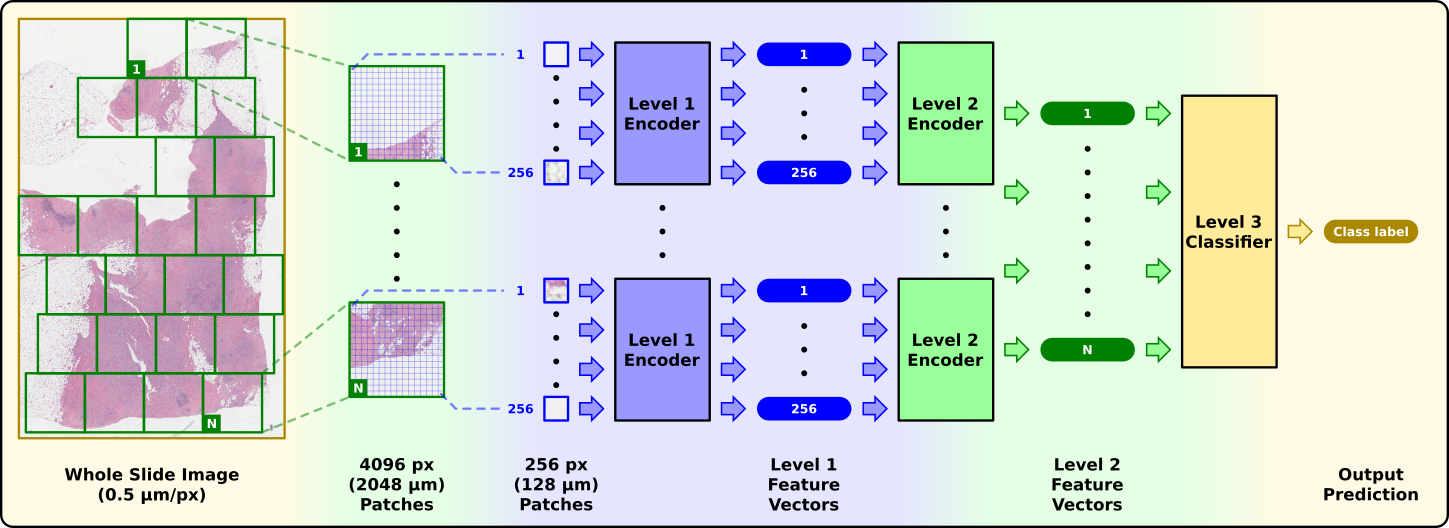}
    \caption{An overview of the Hierarchical Image Pyramid Transformer (HIPT) framework\cite{hipt}. The level 1 transformer encoder is first used to encode each $\SI{256}{\px} \times \SI{256}{\px}$ patch into a level 1 feature vector. Next, a level 2 transformer encoder merges all level 1 feature vectors corresponding to the same $\SI{4096}{\px} \times \SI{4096}{\px}$ patch into a single level 2 feature vector. Finally, the level 3 transformer-based classifier takes all of the level 2 feature vectors together to compute an output class label.  }
    \label{fig:hipt_overview}
\end{figure*}

One of the most successful recent papers that incorporates global information in a very direct way is the Hierarchical Image Pyramid Transformer (HIPT) framework \cite{hipt}. Figure \ref{fig:hipt_overview} shows an overview of how HIPT first applies self-supervised learning to acquire a 384-dimensional embedding for each $\SI{256}{\px} \times \SI{256}{\px}$ level 1 image patch, which are called level 1 feature vectors. Self-supervised learning is applied again at level 2 to acquire a 192-dimensional embedding for each $\SI{4096}{\px} \times \SI{4096}{\px}$ level 2 patch. Finally, all level 2 feature vectors are fed into a single level 3 transformer to make a prediction at the whole slide level. This approach progressively constructs a more global view of the WSI, allowing a hierarchy of transformer models to analyse the global structure. 

In this paper we use the HIPT framework as the basis for our investigation of how important global structure information and self-supervised pre-training are for making good predictions at the WSI level. We do this by systematically stripping global structure information away from HIPT in two ways: 1) reducing the complexity of global structure processing, and 2) reducing the influence of pre-training on global structure. After measuring how this impacts performance, we observe that incorporating more pre-training and more global information does not necessarily give the best accuracy. In fact, incorporating no global structure, a very simple multi-instance learning approach using just level 1 patches can achieve very competitive results. 

Table \ref{tab:overall_results} summarises our key findings. The results are averaged across 4 WSI datasets (CAMELYON16, TCGA-BRCA subtyping, NSCLC subtyping, and RCC subtyping). The columns show varying amounts of level 2 pre-training and rows show varying amounts of global structure. A surprising result is that the very simple max pooling based multi-instance learning (Max-MIL) algorithm~\cite{CLAM} (essentially just using the single most confident level 1 patch prediction) can outperform models that incorporate the most global structure. The results also show that using a pre-trained feature extractor at level 2 does not provide a noticeable benefit. Finally, we find that using a shallow transformer to encode level 2 features (medium global structure) performs the best.

The choice of data used to pre-train the level 1 feature extractor was found to have the biggest impact on overall performance. In our experiments, we used the DINO~\cite{dino} self-supervised feature extractor on combined datasets of varying size. The results show that features learnt from a large collection of 33 cancers performed much worse than features learnt from a smaller subset of 7 cancers, or even learning from only the single target cancer. This may be attributable to the combination of two factors: 1) the large number of patches (e.g.  an average of around 13,258 patches per image) in each WSI being sufficient for learning low-level features, and 2) a large number of different cancer types causing greater divergence between pre-training and the downstream task. For example the ImageNet 1K dataset has size 133GB, which is less than a third of the size of the TCGA-BRCA breast cancer dataset size (480GB), hence WSIs from a single cancer dataset may be enough to learn good discriminative features. Learning from a broader range of cancers may result in reserving precious regions of the embedding space for representations that are not used for the downstream task.

Extensive experiments reveal a simple recipe for modifying HIPT's level 2 encoder to use a shallow transformer (without pre-training) and using a level 1 encoder trained on a smaller, more focused set of 7 cancers (including the target cancer). We call this approach HIPT with local emphasis (\textit{HIPTLE}), and find that it consistently outperforms existing algorithms for all WSI classification and survival prediction tasks tested. The reduced depth of the level 2 encoder allows the final classification module to access the important level 1 information more easily while still being able to use some global context information.

In summary, we make the following key findings in our investigation into what matters for achieving good performance for weakly supervised WSI classification:

\begin{enumerate}
    \item Incorporating global structure information has limited benefits to performance.
    \item The single most significant factor in achieving good performance is the data used to pre-train the level 1 feature extractor.
    \item Pre-training level 1 features on WSIs from a larger range of cancers performs significantly worse than using a smaller set of cancers or even just the target cancer alone.
    \item A very simple max pooling based MIL algorithm that incorporates no global structure information when given high quality pre-trained features can perform similarly to complex state-of-the-art methods.
    \item A modified version of HIPT called HIPTLE consistently outperforms all other algorithms for all WSI classification and survival prediction tasks tested.
   
\end{enumerate}

\begin{table*}[ht]
\centering
\setcellgapes{0.4ex}\makegapedcells
{\footnotesize
\begin{tabular}{lccc}
 \toprule

 & \makecell{Most level 2 pre-training \\ (frozen weights)} & \makecell{Medium level 2 pre-training \\ (fine-tuned weights)} & \makecell{No level 2 pre-training \\ (random initialisation)} \\
\midrule

\makecell{Most global structure \\ (HIPT~\cite{hipt})} & 0.845  & 0.937  & 0.936  \\

\makecell{Medium global structure \\ (HIPTLE)} & 0.872  & 0.954 & \textbf{0.959}   \\

\makecell{No global structure \\ (Max-MIL~\cite{CLAM})} & N/A  &  N/A   &  0.940  \\
\bottomrule
\end{tabular}
}
\caption{Average AUC results across four different public datasets: CAMELYON16 metastases classification, TCGA-BRCA subtyping, NSCLC subtyping, and RCC subtyping. The highest AUC result is highlighted using bold font.  
}
\label{tab:overall_results}
\end{table*}

\section{Related Works}
\label{sec:related_works}
\subsection{Multi-instance Learning}

The majority of existing work in weakly supervised WSI classification takes the multi-instance learning (MIL) approach, where the WSI is represented by a bag of patch instances created by dividing the large WSI into many much smaller tiles. We have identified three main MIL sub-categories, which we refer to as \textit{instance-level simple aggregation} (IL-SA), \textit{instance-level machine learning aggregation} (IL-MLA), and \textit{embedding-level machine learning aggregation} (EL-MLA).

Instance-level simple aggregation (IL-SA) approaches produce separate class label predictions for each patch in the WSI, then apply a simple aggregation function to arrive at the final prediction. Example aggregation functions include taking the maximum probability \cite{campanella2018terabyte}, averaging the probabilities \cite{coudray2018classification}, or counting the percentage of patches predicted to be positive \cite{coudray2018classification}. However, both averaging and using the maximum probability have their different problems. Using the maximum probability can result in many false positives since a single misclassification can change the predicted class \cite{campanella2019clinical-NatureMed}. Averaging suffers from the problem that generally positive regions only occupy small portions of tissue (e.g. less than 20\%), and therefore the vast negative regions overwhelm the positive regions.

Instance-level machine learning  aggregation (IL-MLA) approaches overcome the aforementioned problems of IL-SA by using a machine learning model for more sophisticated combining of per-patch predictions. For example, Hou et al.~\cite{hou2016patch} use logistic regression to combine the instance level predictions. Wang et al.~\cite{wang2016deep} first produce tumor probability heatmaps from the patch level deep learning classifier and then extract geometrical features from the heatmaps. Next they feed the extracted geometrical features into a random forest classifier to make the WSI level predictions. Similarly, Campanella et al.~\cite{campanella2019clinical-NatureMed} train a random forest algorithm on manually engineered features extracted from the patch level heatmaps. These methods use handcrafted features on heatmaps to capture high level spatial structure information from the WSIs.  

Embedding-level machine learning  aggregation  (EL-MLA) approaches generate an embedding (feature vector) for each patch (instance) and then use a machine learning model to combine the instances to arrive at a prediction. In contrast to IL-SA and IL-MLA, this approach allows the model to consider features from the entire WSI when attributing importance to each instance. By leveraging these embeddings, the model can effectively capture the underlying relationships and interactions among instances, leading to more accurate and robust predictions in multi-instance learning tasks. A famous work in this area is the attention based multiple instance learning (ABMIL) paper~\cite{ABMIL}, where an MLP with attention weights is used to automatically learn the importance of each instance for predicting the final slide level binary class label. CLAM~\cite{CLAM} extends this idea to multi-class classification by using multiple attention branches, one for each class. Zhang et al.~\cite{zhang2022dtfd} developed a two-step EL-MLA approach which first randomly samples patches in a WSI to create pseudo bags of patches. They then use an attention-based model to distill the most predictive patches from each pseudo bag and feed those into a second attention based model to make the final prediction.

\subsection{WSI classification incorporating global structure information}

None of the approaches presented in the previous section incorporate global structure information, with the exception of the IL-MLA methods that perform analysis on heatmaps generated from patch level predictions. In this section we focus on techniques that incorporate global structure information when classifying WSIs. These methods all take the EL-MLA approach in the sense that they first use a pre-trained encoder to embed each patch as a feature vector and then train various kinds of models on top of these feature vectors.

One way to capture global structure information is to apply graph convolutional networks (GCNs) on embedded patches of WSIs. Due to the large number of patches, most approaches \cite{adnan2020graph,guan2022node} sample representative patches and then connect the patches using GCNs. Adnan et al.~\cite{adnan2020graph} use a fully connected graph, which essentially means the spatial proximity information is discarded. Guan et al.~\cite{guan2022node} use two levels of graphs. The first level connects patches with similar appearance and the second connects the local graphs using a global graph. This approach does not use spatial location information but instead uses appearance information to determine graph connectivity. 

Another way of incorporating higher level structure information is to ingest embeddings from mixed resolution patches \cite{stegmuller2023scorenet,li2021dual} (e.g. 5X and 20X magnification patches). These approaches capture high level structure by downsampling large image patches into smaller patches (essentially averaging nearby pixels) and then learning embeddings from them using self-supervised learning. A potential drawback of this simple way of compressing high resolution patches is that important low-level features which are critical for making correct predictions may be lost during the averaging of pixel values. In contrast, more recent methods \cite{hipt,adnan2020graph,guan2022node,shao2021transmil,vu2023handcrafted} reduce the dimensionality of patches in more intelligent ways, utilising transformer encoders trained using self-supervised learning.

Some methods~\cite{shao2021transmil,vu2023handcrafted} use transformer self attention layers to capture global structure information. These methods represent patches as tokens with embedded position information. The tokens are then passed into a self attention layer. This allows the model to incorporate global spatial relationship information when making WSI predictions. However, these papers incorporate the position information using a single flat self attention layer. In contrast, the HIPT framework \cite{hipt} takes a hierarchical approach where multiple different transformer models are used to incorporate increasingly higher level structure information. This then opens up the possibility to learn pre-trained features via self supervision at higher levels of the hierarchy (embeddings representing $\SI{4096}{\px} \times \SI{4096}{\px}$ patches instead of $\SI{256}{\px} \times \SI{256}{\px}$ patches).

\section{How much global structure is required?}
\label{sec:vary_global_struct}

Many recent successful WSI classification methods focus on finding the best way to incorporate global information~\cite{hipt,adnan2020graph,guan2022node,shao2021transmil,vu2023handcrafted}. Other recent works did not use any global information at all, treating the WSI as a bag of patches instead~\cite{zhang2022dtfd,ABMIL, campanella2019clinical-NatureMed,CLAM}. Pathologists normally work by first using a zoomed-out view of the WSI for an overview of the tissue sample, and then zoom in to areas of interest (high-power fields) for more detailed analysis. The cell level information contained in high-power fields is highly relevant for both cancer sub-typing and survival prediction. We suspect there is a trade-off between focusing on the global information and focusing on the low level cell information. Methods like HIPT---which has a deep level 2 encoder for processing global structure---have a large degree of separation between the final classifier and the low level cell information at level 1 of the hierarchy. This makes the model less sensitive to cell level information when making predictions. In contrast, methods that treat the WSI as a collection of individual small patches have a much flatter model structure which allows the training signal (class label) to reach the cell level much easier. 

In this paper we study the importance of global structure information for making accurate predictions on 4 different WSI classification tasks (CAMELYON16 metastases prediction, breast cancer sub-typing, kidney cancer sub-typing, and lung cancer sub-typing). To do this we consider three model configurations that capture different levels of global structure information, effectively varying the ``distance'' (in terms of the number of layers) between level 1 feature vectors and the classification output. At one extreme, the final classifier sees more global information at the cost of being further away from the input. At the other extreme, the final classifier is closer to the input but does not see as much global information. Comparisons are made based on the same level 1 encoder layer---a ViT-s \cite{VIT} model pre-trained using DINO\cite{dino} on the same set of WSIs.

\begin{figure*}[htbp]
    \begin{subfigure}{0.3\textwidth}
        \includegraphics[width=\textwidth]{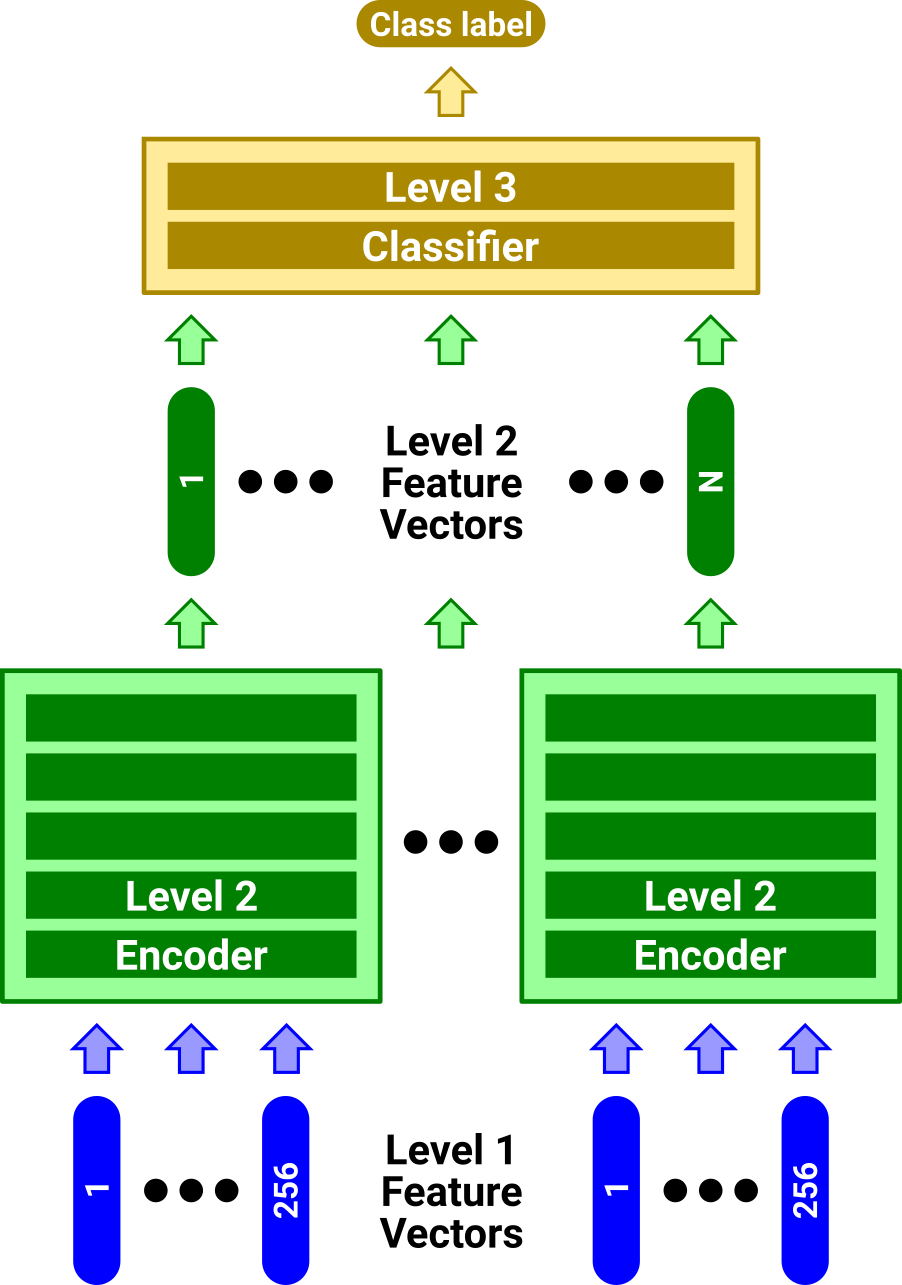}
        \caption{Most global structure (HIPT)}
        \label{fig:most_global_structure}
    \end{subfigure}
    \hfill
    \begin{subfigure}{0.3\textwidth}
        \includegraphics[width=\textwidth]{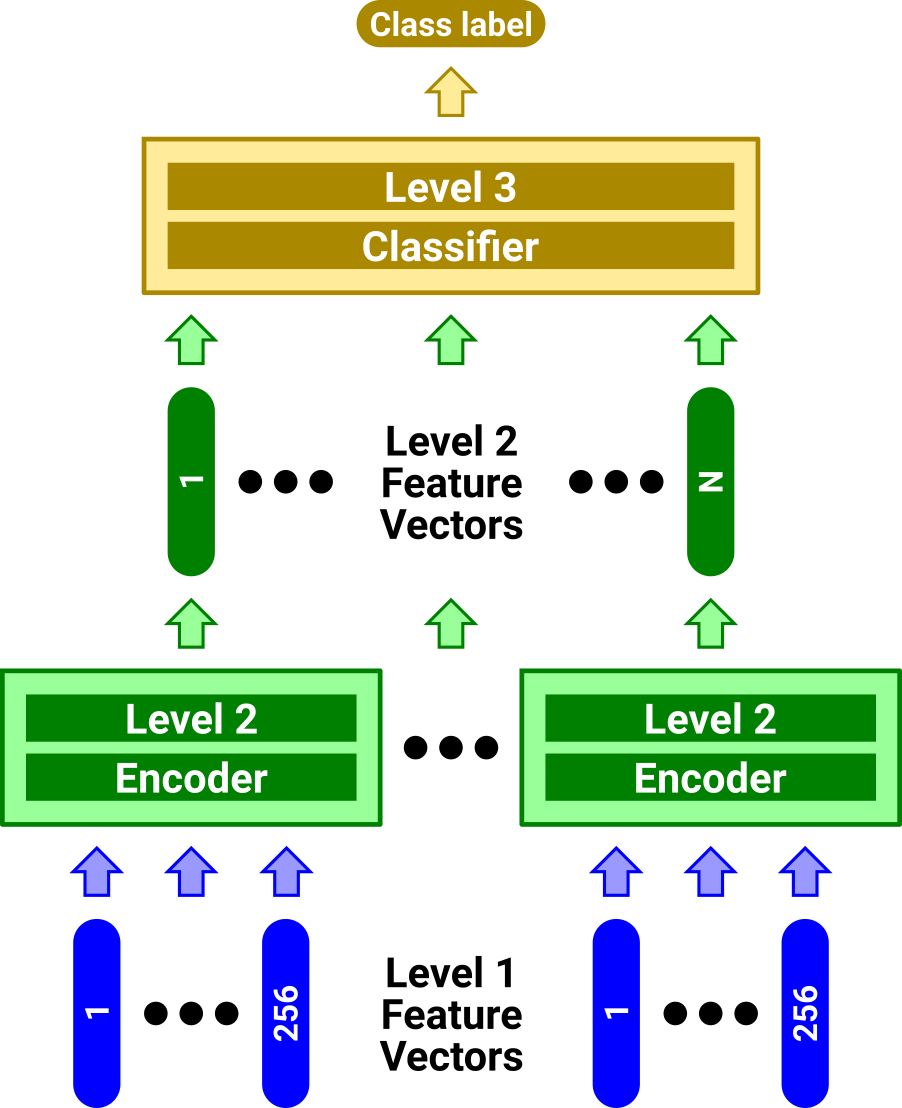}
        \caption{Medium global structure (HIPTLE)}
        \label{fig:medium_global_structure}
    \end{subfigure}
    \hfill
    \begin{subfigure}{0.3\textwidth}
        \includegraphics[width=\textwidth]{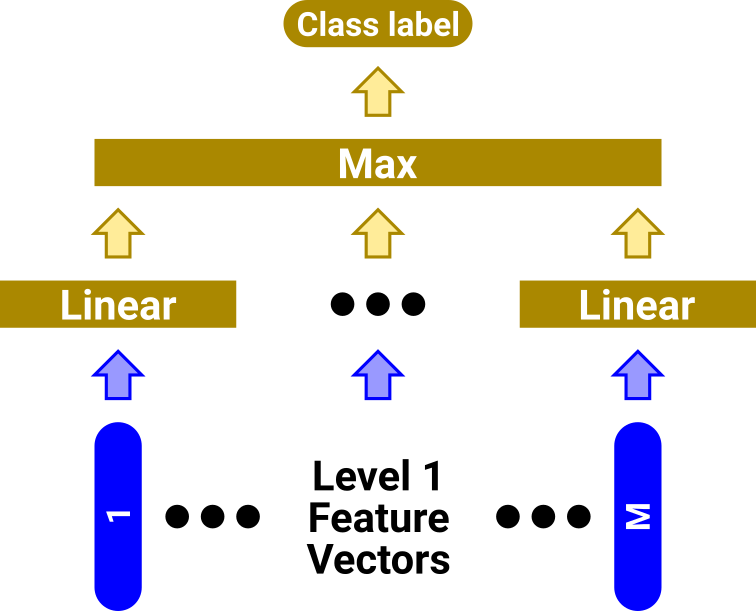}
         \caption{No global structure (Max-MIL)}
         \label{fig:no_global_structure}
    \end{subfigure}
    \caption{Models with three different levels of global structure used for WSI classification predictions. (\subref{fig:most_global_structure}) Most global structure corresponds to the original HIPT framework, which has a deep transformer as its level 2 encoder. (\subref{fig:medium_global_structure}) Medium global structure replaces the level 2 encoder of HIPT with a shallower 2 layer transformer model. (\subref{fig:no_global_structure}) No global structure (Max-MIL) uses a simple max operator to aggregate the individual contributions from each patch without incorporating any position or structure information. All models use the same pre-trained level 1 encoder (not shown) to produce level 1 feature vectors.}
    \label{fig:varying_global_structure}
\end{figure*}

Figure \ref{fig:varying_global_structure} shows the three different model designs that we tested for varying amounts of global structure. The first design (Figure \ref{fig:most_global_structure}) incorporates the most global structure information. It corresponds to the original HIPT framework setup \cite{hipt}, as illustrated in Figure \ref{fig:hipt_overview}. In this model design three levels of transformers are used to arrive at the final prediction. The level 2 encoder combines the 384D vector representing the level 1 patches using position information to give the model a complete structural view of large $\SI{4096}{\px} \times \SI{4096}{\px}$ patches. At level 2, the model should have enough context to be able to analyse tissue architecture information such as invasive fronts and neoplastic structures. Finally, the 192D level 2 feature vectors are passed into the final transformer classifier to arrive at a prediction for the WSI. Although this approach may allow the model to see more global structure in the WSI, the downside is that the many layers of high-level processing make it harder for the classifier to incorporate important cell level information.

To make the low level cell information more accessible to the final classification layer we replaced the deep 6 layer transformer network using 6 heads with a shallower 2 layer transformer network using 3 heads. This shallower level 2 encoder allows the information from the level 1 feature vectors to more easily flow to the final classification transformer. 

\begin{figure}[t]
    \centering
    \includegraphics[width=6cm]{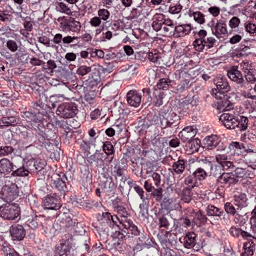}
    \caption{A $\SI{256}{\px} \times \SI{256}{\px}$ image patch at 20X magnification from the TCGA-BRCA dataset. You can see the cells and their spatial arrangement clearly within the tissue micro-environment.  }
    \label{fig:256crop}
\end{figure}

Finally, we used the simple multi-instance learning approach that treats each level 1 feature vector as a separate instance, and each instance is fed into an MLP to separately predict the slide label. For binary classification the patch with the highest predicted probability for the positive class is selected to decide the predicted class for the entire slide as well as gradient signals during training. To handle multi-class classification the MLP is modified to predict multiple classes. The patch with the highest single class probability score across all classes is used to make the slide-level label prediction.  We call this the Max-MIL approach and we take the implementation from CLAM~\cite{CLAM}. This approach does not incorporate any global structural information beyond the $\SI{256}{\px} \times \SI{256}{\px}$ patch, and therefore the model is not able to learn spatial patterns that are larger than this. Figure \ref{fig:256crop} shows an example $\SI{256}{\px} \times \SI{256}{\px}$ patch at 20X magnification. The type of cells, local spatial arrangement of cells, and tissue type information are all visible at this magnification level.

\section{Varying the amount of pre-training}
\label{sec:vary_pre_training}

\begin{figure*}[htbp]
    \captionsetup[subfigure]{justification=centering}
    \begin{subfigure}{0.3\textwidth}
        \includegraphics[width=\textwidth]{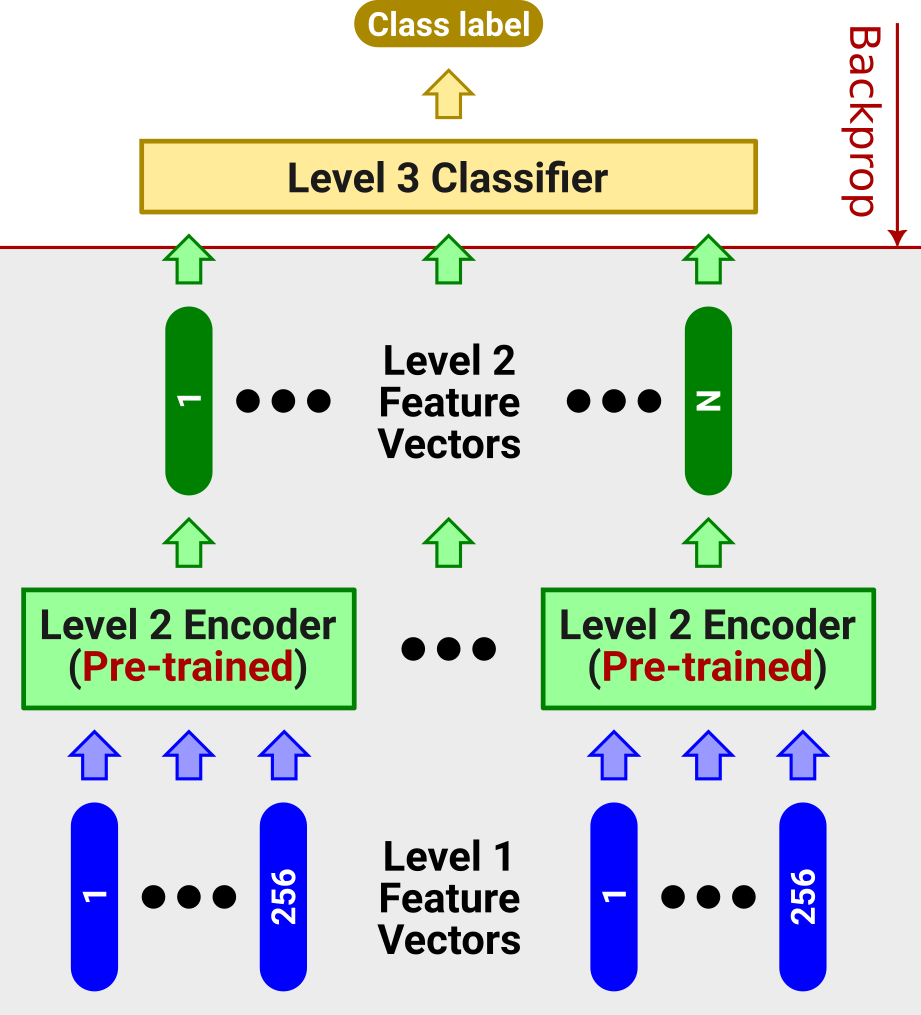}
        \caption{Most level 2 pre-training \\(frozen level 2)}
        \label{fig:image1}
    \end{subfigure}
    \hfill
    \begin{subfigure}{0.3\textwidth}
        \includegraphics[width=\textwidth]{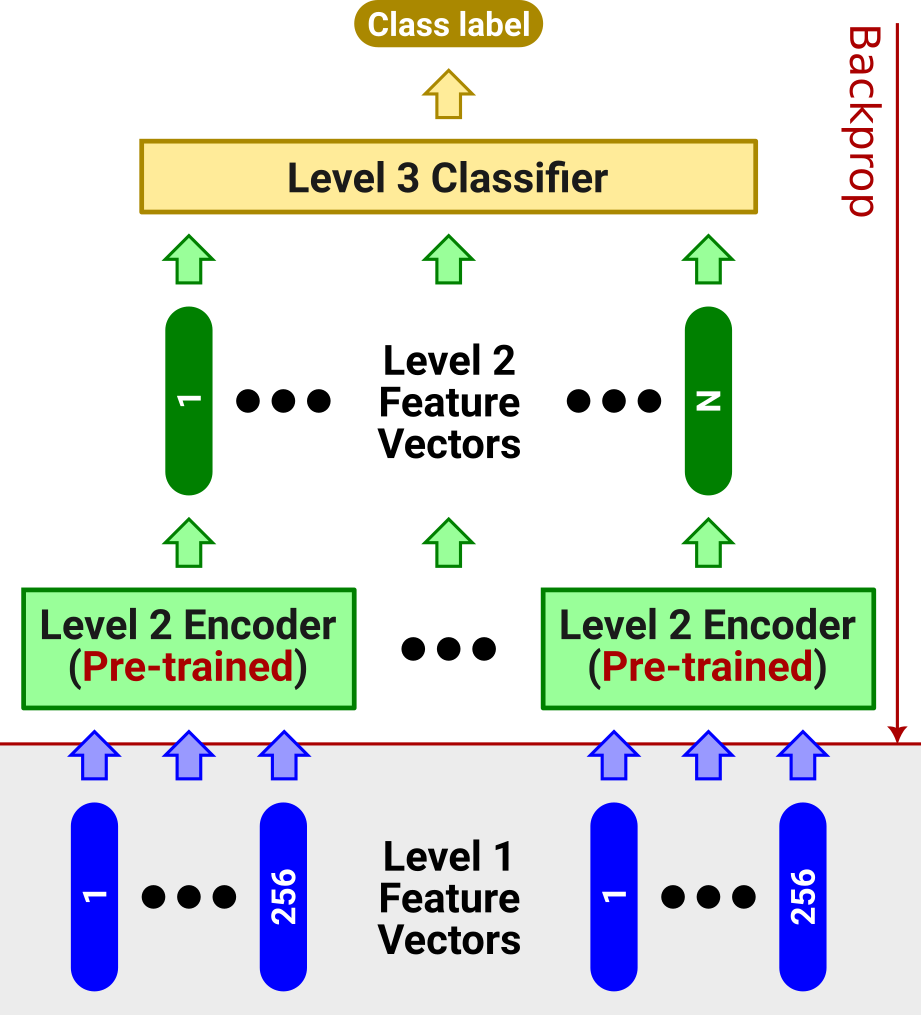}
        \caption{Medium level 2 pre-training \\(fine-tuned level 2)}
        \label{fig:image2}
    \end{subfigure}
    \hfill
    \begin{subfigure}{0.3\textwidth}
        \includegraphics[width=\textwidth]{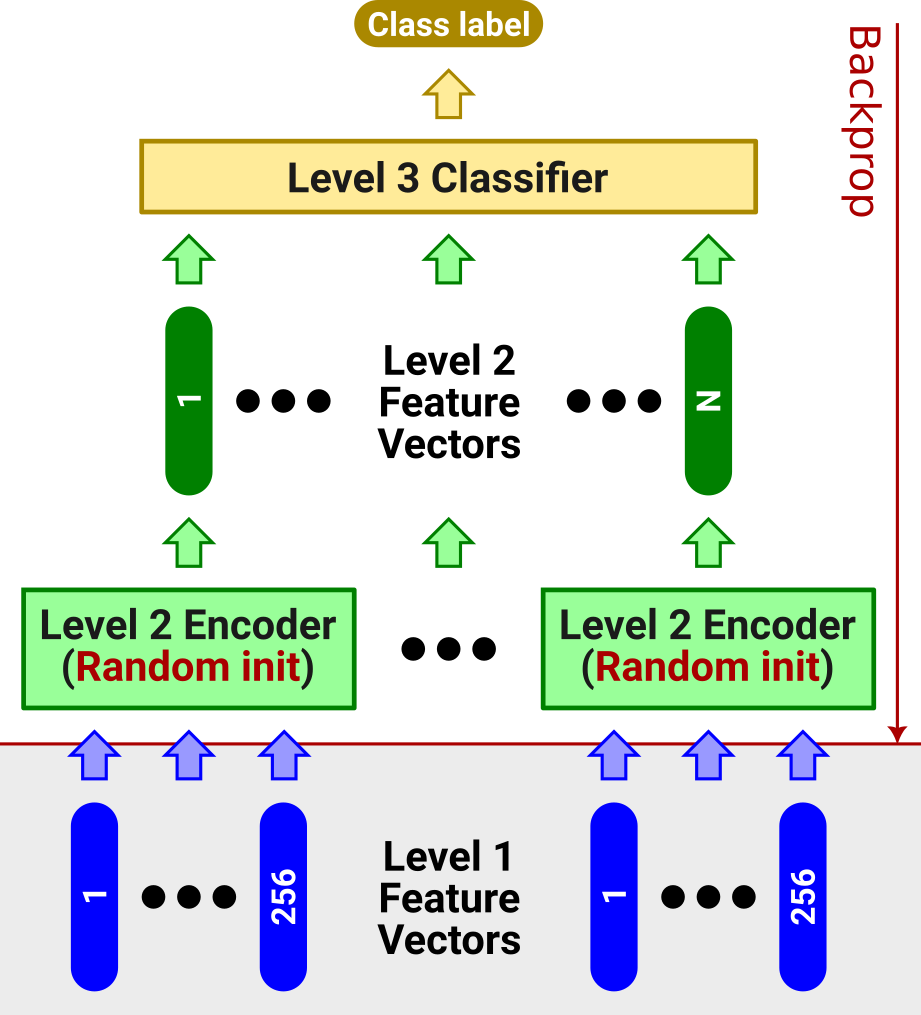}
         \caption{No level 2 pre-training \\(train randomly initialised level 2)}
         \label{fig:image3}
    \end{subfigure}
    \caption{Three different training configurations with varying levels of pre-trained weight utilization in the level 2 encoder. All models use the same pre-trained level 1 encoder (not shown) to produce level 1 feature vectors.}
    \label{fig:varying_pre_training}
\end{figure*}

The HIPT framework \cite{hipt} advocates pre-training both the level 1 and level 2 encoders on large unlabelled datasets. Using heavily pre-trained level 2 encoders means the models start with good high level feature extractors that span large $\SI{4096}{\px} \times \SI{4096}{\px}$ patches. It also opens up the possibility of freezing those weights and restricting optimisation on the downstream task to the level 3 classifier only. Intuitively this should have two key benefits. Firstly, the model should be able to find useful global patterns during pre-training and reuse these for the downstream task, thus resulting in better performance compared to random initialisation. Secondly, freezing the level 2 encoder weights should reduce the likelihood of overfitting on the downstream task, since the model is more constrained.

Given that the idea of pre-training at the high $\SI{4096}{\px} \times \SI{4096}{\px}$ patch level is relatively new, we wanted to empirically evaluate how beneficial such pre-training is in practice. To do this, we tested three different training configurations which vary in the amount of level 2 pre-training. Note that fine-tuning the level 1 encoder is not feasible due to memory constraints of current computing resources.  

The three training configurations are shown in Figure \ref{fig:varying_pre_training}. The first training configuration has the most level 2 pre-training, and is the best-performing configuration from the HIPT framework. The pre-trained level 1 and level 2 features are frozen and we only train the level 3 classifier on the downstream task. This is similar to the linear probing method for transfer learning where only the final linear head is trained on the downstream task. In general this approach should give the model the least chance to overfit to the training set since most of the hierarchical model (both level 1 and 2) is frozen.

The second training configuration, ``medium level 2 pre-training'', fine-tunes the pre-trained level 2 encoder parameters while training on the downstream task. This configuration makes use of the pre-training to ensure the model starts off with good initial weights before it is fine-tuned for the target task.

The third training configuration does not use any level 2 pre-training, and instead randomly initialises level 2 weights before training on the downstream task. In theory this configuration has the most opportunity to overfit the training data since the level 2 weights are adjusted solely based on the down stream task training data.

\section{Experimental setup}
In this section we describe the datasets, data preprocessing, metrics, train/test splits, and training setup used to conduct our experiments.

\subsection{Datasets}
\label{sec:datasets}
We used the same datasets as \cite{hipt}, but expanded our evaluation to also include the CAMELYON16 dataset~\cite{bejnordi2017diagnostic}. So we used the following public datasets: TCGA-BRCA; TCGA-LUAD; TCGA-LUSC; TCGA-KIRC; TCGA-KIRP; and CAMELYON16. Using the TCGA-BRCA dataset we performed Invasive Ductal (IDC) versus Invasive Lobular Carcinoma (ILC) subtyping with a total of 937 WSIs. We combined TCGA-LUAD and TCGA-LUSC datasets to perform Lung Adenocarcinoma (LUAD) versus Lung Squamous Cell Carcinoma (LUSC) in Non-Small Cell Lung Carcinoma (NSCLC) subtyping with a total of 958 WSIs. We combined TCGA KIRC and TCGA KIRP to perform Clear Cell, Papillary, and Chromophobe Renal Cell Carcinoma (CCRCC vs. PRCC vs. CHRCC) subtyping with a total of 931 WSIs. Finally we performed metastases binary classification using the CAMELYON16 dataset which consisted of 270 training images and 129 validation images.

Apart from Section \ref{sec:vary_level_1_encoder} (where we varied the pre-training dataset used), all our experiments used the following 7 cancer datasets to pre-train the level 1 encoder: CPTAC-COAD, PAIP2019\cite{kim2021paip}, TCGA BRCA, TCGA LUAD, TCGA LUSC, TCGA KIRC and TCGA KIRP.

\subsection{Data preprocessing}

The WSIs are rescaled to a consistent base magnification level of 0.5 microns/pixel. Macenko normalisation~\cite{macenko} is applied to the WSIs to achieve a canonical colouring of haematoxylin and eosin stains. Using calculated tissue masks for separating tissue from the slide background, we extracted $\SI{256}{\px} \times \SI{256}{\px}$ patches that have at least 75\% foreground pixels. These patches were used for training the level 1 encoder. The level 2 encoder was trained on $\SI{4096}{\px} \times \SI{4096}{\px}$ patches with at least 40\% foreground pixels. These same level 2 patches were also used for the downstream tasks. We set this threshold lower for the CAMELYON16 dataset (to 20\% foreground pixels), since the task is to find very small cancerous cells in the WSIs and setting a high foreground threshold could lose important information.

\subsection{Cross Validation and metrics}
We followed the experimental protocol of \cite{hipt}, performing 10 fold cross validation on all experiments involving TCGA datasets. We used the same cross validation splits as those used in \cite{hipt}. For the CAMELYON16 dataset we used the train and validation split provided by the challenge as our train and validation splits. We used the AUC metric for all binary classification tasks including cancer subtyping (TCGA) and classification of metastases (CAMELYON16). For RCC subtyping---which has three classes---we report macro-averaged AUC. 

\subsection{Training setup}

We pre-trained on the 7-cancer dataset containing 3565 WSIs, which consisted of 39,660,927 level 1 patches and 200,966 level 2 patches. We trained the level 1 encoder for 1600 epochs using the ViT-s \cite{VIT} architecture and AdamW \cite{adamw} optimizer with a base learning rate of 0.0005 and a batch size of 32. The first 10 epochs were used to warm up to the base learning rate followed by a cosine schedule decay. Due to the massive number of level 1 patches, we defined an ``epoch'' to be smaller than a full pass through the dataset. By our definition of an epoch, the model will see a total of $2^{16}=65536$ training examples randomly sampled from the entire dataset.

The level 2 encoder was trained with similar configuration settings using the standard definition of an epoch (one full pass through the dataset). The model with ViT-xs architecture was trained for 800 epochs using the level 1 feature vectors.

For most finetuning experiments, we trained for 20 epochs with the Adam \cite{kingma2014adam} optimizer, batch size of 1 and a learning rate of 0.0001. The metastases prediction task on the CAMELYON16 dataset was finetuned for 100 epochs as an exception.

\section{Experimental Results}
In this section we present the results from extensive experiments we have performed to test what really matters for determining the performance of WSI classification models. The factors we tested include the following: the amount of global information the model incorporates; the amount of level 2 pre-training used; the number of different cancer datasets used to pre-train the level 1 encoder; and the amount of training data used. Finally we tested the performance of the models for survival prediction.

In our experiments we found that there was one model configuration which almost always gave the best results. We call this configuration \textit{HIPT with local emphasis} (HIPTLE). HIPTLE uses the medium global structure model (see Section \ref{sec:vary_global_struct} for details), with no level 2 pre-training and uses a level 1 encoder trained using the 7 cancers listed at the end of Section \ref{sec:datasets}. Many of the experiments below will include results for HIPTLE. 

\subsection{Varying the amount of pre-training and global structure}
\label{sec:vary_pre-traing_global}
\begin{table*}[ht]
\centering
\setlength{\tabcolsep}{4pt}
{\scriptsize
\begin{tabular}{lccc L{0cm} ccc}
 \toprule
 & \multicolumn{3}{c}{CAMELYON16 metastases} & \phantom{abc} & \multicolumn{3}{c}{BRCA subtyping} \\
 \cmidrule{2-4} \cmidrule{6-8}
 & Most L2 PT & Med L2 PT & No L2 PT && Most L2 PT & Med L2 PT & No L2 PT \\
 \midrule
Most global structure & 0.564 & 0.951& 0.931 && 0.800 $\pm$ 0.072 & 0.884 $\pm$ 0.068 & 0.878 $\pm$ 0.053 \\
Med global structure   & 0.666 & \bf{0.964} & 0.960 && 0.882 $\pm$ 0.039  & 0.900 $\pm$ 0.036  & \bf{0.916 $\pm$ 0.038}  \\
No  global structure  & - & - & 0.952&& - & - & 0.879 $\pm$ 0.0729 \\
\midrule
DTFD-MIL~\cite{zhang2022dtfd}, HIPT~\cite{hipt}  & - & - & 0.945  && 0.874 $\pm$ 0.060 & $0.827 \pm 0.069$ & $0.823 \pm 0.071$ \\
\midrule
\midrule
 & \multicolumn{3}{c}{NSCLC subtyping} & \phantom{abc} & \multicolumn{3}{c}{RCC subtyping} \\
\cmidrule{2-4} \cmidrule{6-8}
 & Most L2 PT & Med L2 PT & No L2 PT  && Most L2 PT & Med L2 PT & No L2 PT \\
\midrule
Most global structure & 0.874 $\pm$ 0.038 & 0.951 $\pm$ 0.020 & 0.953 $\pm$ 0.019 && 0.976 $\pm$ 0.013 & 0.989 $\pm$ 0.009 &  0.985 $\pm$ 0.010 \\
Med global structure   & 0.950 $\pm$ 0.020 & 0.960 $\pm$ 0.015 & \bf{0.965 $\pm$ 0.013} && 0.991 $\pm$ 0.006 & 0.993 $\pm$ 0.005 & \bf{0.993 $\pm$ 0.004} \\
No  global structure  & - & - &  0.940 $\pm$ 0.028&& - & -  &  0.991 $\pm$ 0.005\\
\midrule
HIPT~\cite{hipt}   & 0.952 $\pm$ 0.021 & $0.820 \pm 0.047$ & $0.786 \pm 0.096$  && $0.980 \pm 0.013$ & $0.956 \pm 0.013$ & $0.956 \pm 0.016$\\
\bottomrule
\end{tabular}
}
\caption{AUC results from four different public datasets: CAMELYON16 metastases classification, TCGA-BRCA subtyping, NSCLC subtyping, and RCC subtyping. The comparison algorithm used for the CAMELYON16 dataset was DTFD-MIL\cite{zhang2022dtfd} and HIPT\cite{hipt} was used for all other datasets. The highest AUC result for each dataset is highlighted using bold font. Note most/med/no L2 PT, refers to most/med/no level 2 pre-training. }
\label{tab:individual_results}
\end{table*}

In the introduction we showed the overall results across 4 datasets when both the influence of pre-training and the amount of model capacity dedicated to global structure was varied. Table \ref{tab:individual_results} shows a more detailed breakdown of these results, considering each dataset individually. We used the definitions of most/medium/no level 2 pre-training from Section \ref{fig:varying_global_structure} and most/medium/no global structure from Section \ref{sec:vary_pre_training}. The results show that the HIPTLE configuration of using a medium amount of global structure and fine-tuning the level 2 encoder (either Med L2 PT or No L2 PT) gives the best performance for all datasets. Starting with random weights for the level 2 encoder (No L2 PT) or with pre-trained  weights (Med L2 PT) for the level 2 encoder does not make much difference. This shows level 2 encoder pre-training is not effective.

The ``no global structure" configuration was a surprisingly strong performer, achieving results that were close to the best result for three of the WSI classification tasks (CAMELYON16 metastases, NSCLC subtyping, and RCC subtyping). This suggests that most of the information needed to successfully classify each WSI resides at the low $\SI{256}{\px} \times \SI{256}{\px}$ patch level (level 1 encoder), where cell type, cell density, and tissue type information can be determined. Put another way, pre-training the level 2 encoder is less important for accurate predictions than how well valuable information from the level 1 layer is transmitted to the final layer during supervised training (either by omitting global structure or by fine-tuning the level 2 encoder).

The results show that medium global structure models always outperform the most global structure models for any pre-training configuration. This shows the importance of not making the models too deep. As mentioned above it seems the low level cell information is really useful for the final prediction and so using fewer layers before the level 3 classification module allows the low level information to be passed to the final prediction layers more easily, resulting in better performance.

It is also important to note the method using medium global structure while fine-tuning the level 2 encoder outperforms DTFD-MIL~\cite{zhang2022dtfd} (for CAMELYON16) and HIPT~\cite{hipt} (for BRCA, NSCLC and RCC subtyping). This shows our models give very strong performance when compared with existing state-of-the-art methods.

\subsection{Varying data used to pre-train level 1 encoder}
\label{sec:vary_level_1_encoder}

\begin{table*}[ht]
\centering
\setlength{\tabcolsep}{4pt}
{\scriptsize
\begin{tabular}{cccccc}
\toprule
 Pre-training dataset& CAMELYON16 & BRCA subtyping & NSCLC subtyping & RCC subtyping & Average \\
 \midrule
 33 cancers & 0.763  & 0.748 ± 0.090 & 0.886 ± 0.027 & 0.951 ± 0.015 & 0.840 \\
  7 cancers &  0.952 & 0.879 $\pm$ 0.0729 &  0.940 $\pm$ 0.028 & {\bf 0.991 $\pm$ 0.005} & 0.941\\
 single cancer & {\bf 0.963} & {\bf 0.888 ± 0.067}  &  {\bf 0.947 ± 0.026} & 0.981 ± 0.013 & {\bf 0.945} \\
 ImageNet & 0.800 & 0.850 ± 0.083 & 0.907 ± 0.026 & 0.943 ± 0.021 & 0.875\\
 \bottomrule
\end{tabular}
}
\caption{AUC results for the no global structure model (Max-MIL) when varying the dataset used for pre-training the level 1 encoder. The 7 cancers dataset is the default dataset used to train level 1 encoders for all the experiments (see Section \ref{sec:datasets}). The 33 cancers pre-training results was using the pre-trained level 1 encoder from HIPT\cite{hipt} which was trained on 33 cancers. For single cancer results we pre-trained the level 1 encoder using the same dataset as that used for the downstream WSI classification task. The best result for each dataset is highlighted in bold font. }
\label{tab:vary_level_1_MIL}
\end{table*}

\begin{table*}[ht]
\centering
{\scriptsize
\begin{tabular}{cccccc}
\toprule
 Pre-training dataset& CAMELYON16 & BRCA subtyping & NSCLC subtyping & RCC subtyping & Average \\
 \midrule
 33 cancers &  0.652 &  0.855  $\pm$ 0.073 &  0.976  $\pm$ 0.009 & 0.924  $\pm$ 0.027  &  0.852 \\
  7 cancers & {\bf 0.960}   & {\bf 0.916  $\pm$ 0.038}  & {\bf 0.993  $\pm$ 0.004}  & {\bf 0.965  $\pm$ 0.013} &  {\bf 0.959}\\
 single cancer &  0.960 & 0.911  $\pm$ 0.049  & 0.987  $\pm$ 0.008  &  0.961  $\pm$ 0.029 & 0.955\\
 ImageNet & 0.813  &  0.896  $\pm$ 0.054 & 0.984  $\pm$ 0.006  & 0.945  $\pm$ 0.019  &  0.910 \\
 \bottomrule
\end{tabular}
}
\caption{ AUC results for the HIPTLE model when varying the dataset used for pre-training the level 1 encoder. The 7 cancers dataset is the default dataset used to train level 1 encoders for all the experiments (see Section \ref{sec:datasets}). The 33 cancers pre-training results used the pre-trained level 1 encoder from HIPT\cite{hipt} which was trained on 33 cancers. For single cancer results we pre-trained the level 1 encoder using the same dataset as that used for the downstream WSI classification task. The best result for each dataset is highlighted in bold font. }
\label{tab:vary_level_1_med_global}
\end{table*}

The previous experiments established that the level 1 encoder extracted the most valuable information for WSI classification and that the level 2 encoder was comparatively less important. This motivated us to explore pre-training the level 1 encoder using different datasets. For all the experiments we used the DINO\cite{dino} unsupervised training method with the VIT-S \cite{VIT} vision transformer model (the same setup used by the HIPT paper~\cite{hipt}). We show the results for the no global structure (Max-MIL) and HIPTLE models (refer to Section \ref{sec:vary_global_struct}). 

Our results in Table \ref{tab:vary_level_1_MIL} show that pre-training the Max-MIL level 1 encoder on the single cancer that was used for downstream WSI classification leads to the best accuracy. The 7-cancer dataset is a close second, but 33-cancer and ImageNet pre-training perform much worse. We think the reason for this is pre-training on fewer cancers (1 or 7) results in the representation space being better utilised to embed just the features that are found on the these small set of cancers instead of the higher amount of irrelevant features found in the 33-cancer dataset or ImageNet. This means smaller differences in features will be mapped farther away in the representation space. In contrast, the 33-cancer pre-trained level 1 encoders need to reserve representation space for cancers that are not part of the downstream WSI classification task.

The results for the HIPTLE model are shown in Table \ref{tab:vary_level_1_med_global}. The results once again show pre-training on fewer cancer types (1 or 7) works better than 33 cancers or ImageNet pre-training. This is for similar reasons to the results for the Max-MIL model. 

Pre-training on ImageNet gives consistently poor results. This can be explained by the fact that feature extractors trained on ImageNet reserve areas of the representation space for features pertaining to natural images, such as photos of dogs. Whilst some low-level features learned during pre-training can be reused, other features simply never appear in WSIs, and hence that portion of the representation space is wasted. In contrast, pre-training on cancer datasets close to the downstream task makes the most efficient use of the representation space to encode features that are most useful for performing classification on WSIs.

\subsection{Frozen versus fine-tuning L2 encoder}

\begin{table}[ht]
\centering
{\scriptsize
\begin{tabular}{cccccc}
\toprule
 Pre-training dataset & L2 encoder & Test AUC \\
 \midrule
 7 cancers & Fine-tuned & 0.892 $\pm$ 0.039 \\
 7 cancers & Frozen & 0.819 $\pm$ 0.091 \\
 33 cancers & Fine-tuned & 0.866 $\pm$ 0.051 \\
 33 cancers & Frozen & \bf{0.902 $\pm$ 0.058} \\
 \bottomrule
\end{tabular}
}
\caption{Test results for the BRCA subtyping task after HIPT models were trained for 100 epochs. Here we vary the pre-training dataset and whether or not the L2 encoder is frozen during supervised learning.}
\label{tab:frozen_vs_finetuned_l2}
\end{table}

\begin{figure}[t]
    \centering
    \includegraphics[width=7cm]{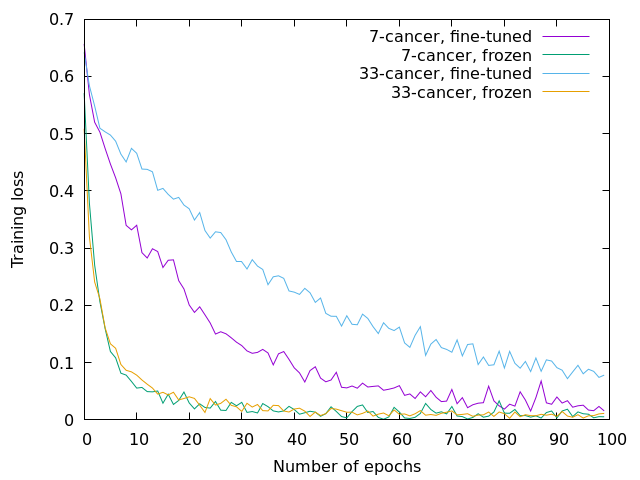}
    \caption{Loss curves from training HIPT on the BRCA subtyping supervised classification task. Regardless of whether pre-training used the 7-cancer or 33-cancer dataset, the HIPT model's training loss improves much quicker with frozen L2 encoder weights}
    \label{fig:train_loss_pretrained_l2}
\end{figure}

The authors of the original HIPT paper found that using a frozen L2 encoder, pre-trained on 33 cancer types, resulted in the best accuracy. In contrast, our experimental results indicate that fine-tuning the L2 encoder (No L2 PT) works best when pre-trained on 7 cancer types (see Table \ref{tab:individual_results}).

The key to understanding this discrepancy is the difference between the two pre-training datasets. As we have already established, 7-cancer data produces better L1 features than 33-cancer data due to better alignment with the downstream task. This is most evident in the Max-MIL results (Table \ref{tab:vary_level_1_MIL}). The reduction in number of examples from the 33-cancer dataset to 7-cancer dataset does not hinder performance, since there are still ample L1 patches in the 7-cancer dataset (39,660,927 patches) to learn good discriminative features. However, this no longer holds true when it comes to pre-training the L2 encoder. Since L2 patches are much larger ($\SI{4096}{\px} \times \SI{4096}{\px}$), there are far fewer examples available for L2 pre-training and dataset size becomes a more critical issue, favouring the 33-cancer data. For the 7-cancer dataset there are only 200,966 L2 patches (99.5\% fewer than L1 patches).

Considering the original HIPT setup we can compare performance on the downstream classification task of BRCA subtyping. The results in Table \ref{tab:frozen_vs_finetuned_l2} show that, when the L2 encoder is frozen, pre-training on 33-cancer data gives better results than 7-cancer data. This is because despite the L1 features from the 7-cancer data having more discriminative power for the downstream task, the relatively small amount of L2 training data leads to overall worse performance. On the other hand, when L2 encoder weights are fine-tuned during supervised training we observe the opposite result. This is because once L2 is fine-tuned then the superior quality of 7 cancer L1 features (recall that the L1 encoder is always frozen) leads to better overall performance.

It takes longer to fine-tune the large L2 encoder of HIPT than it does to keep it frozen during the supervised learning phase (see Figure \ref{fig:train_loss_pretrained_l2} for training loss curves) due to the much larger number of parameters that need to be optimized. After an extended training time of 100 epochs, the best evaluation results are still obtained from the 33-cancer, frozen L2 configuration. However, when the L2 encoder is much shallower (as in HIPTLE, Table \ref{tab:vary_level_1_med_global}), fine-tuning is much more efficient and the strong training signal from supervised training places greater emphasis on having stronger L1 features than stronger L2 pre-training. As a consequence, we find that fine-tuning from 7-cancer data is the best configuration for HIPTLE and hence this is the approach that we will compare with existing methods.

\subsection{Comparison against existing algorithms for WSI classification}
\label{sec:comp_algs_WSI_classification}
\begin{table*}[ht]
\centering
\setlength{\tabcolsep}{4pt}
{\scriptsize
\begin{tabular}{lcccccc}
 \toprule
 & \multicolumn{2}{c}{ BRCA Subtyping } & \multicolumn{2}{c}{ NSCLC Subtyping } &  \multicolumn{2}{c}{ RCC Subtyping} \\
 \cmidrule{2-3} \cmidrule{4-5} \cmidrule{6-7}
 Architecture & 25\% Training & 100\% Training & 25\% Training  & 100\% Training & 25\% Training & 100\% Training \\
 \midrule
CLAM-SB\cite{CLAM} & 0.796 $\pm$ 0.063 &  0.858 $\pm$ 0.067 & 0.852 $\pm$ 0.034 & 0.928 $\pm$ 0.021 & 0.957 $\pm$ 0.012 & 0.973 $\pm$ 0.017  \\
DeepAttnMISL\cite{yao2020whole}  & 0.685 $\pm$ 0.110 & 0.784 $\pm$ 0.061 &   0.663 $\pm$ 0.077 & 0.778 $\pm$ 0.045 &  0.904 $\pm$ 0.024 &  0.943 $\pm$ 0.016\\
GCN-MIL\cite{li2018graph,zhao2020predicting}  & 0.727 $\pm$ 0.076 &  0.840 $\pm$ 0.073 & 0.748 $\pm$ 0.050 & 0.831 $\pm$ 0.034 & 0.923 $\pm$ 0.012 & 0.957 $\pm$ 0.012 \\
DS-MIL\cite{li2021dual}  & 0.760 $\pm$ 0.088 &  0.838 $\pm$ 0.074 & 0.787 $\pm$ 0.073 & 0.920 $\pm$ 0.024 & 0.949 $\pm$ 0.028 & 0.971 $\pm$ 0.016 \\
HIPT\cite{hipt} & 0.821 $\pm$  0.069 & 0.874 $\pm$  0.060 & 0.923 $\pm$  0.020 & 0.952 $\pm$  0.021 & 0.974 $\pm$  0.012 & 0.980 $\pm$  0.013 \\
\makecell[l]{No global structure\\(Max-MIL)~\cite{CLAM}} &  0.828 $\pm$  0.076 &  0.879 $\pm$  0.073 & 0.923 $\pm$  0.032 &  0.94 $\pm$  0.0278 &  0.880 $\pm$  0.0381 & 0.990 $\pm$  0.005 \\
HIPTLE & {\bf 0.864 $\pm$ 0.060} & {\bf 0.916 $\pm$ 0.038} & {\bf 0.943 $\pm$ 0.027} & {\bf 0.965 $\pm$ 0.013} & {\bf 0.989 $\pm$ 0.008} &  {\bf 0.993 $\pm$ 0.004} \\
\bottomrule
\end{tabular}
}
\caption{Experimental results comparing existing WSI classification algorithms against the no global structure (Max-MIL) and HIPTLE configurations reported in this paper. The results for the first 5 algorithms above are taken from the HIPT paper~\cite{hipt}. The best result in each column is highlighted using bold font.}
\label{Table:results_existing_algs}
\end{table*}

In this experiment we compare the no global structure (Max-MIL) and HIPTLE configurations against an array of existing WSI classification algorithms. The results show that HIPTLE significantly outperforms all other algorithms for both 25\% and 100\% training data. This can be attributed to finding the sweet spot in terms of both the degree of pre-training and the amount of global structure. As discussed earlier, a deep model---such as that used in HIPT---can result in the important cell level information being lost before reaching the level 3 classification module. 

The results for the no global structure (Max-MIL) method were similar to HIPT for most of the test configurations despite the fact Max-MIL does not see any global context information. This can be largely attributed to the fact the Max-MIL model used the level 1 encoder that was trained on the 7 cancers instead of the 33 cancers that was used to train HIPT. As we showed in Section \ref{sec:vary_level_1_encoder}, pre-training on the 7 cancers produces better results since it makes better use of the representation space.

\subsection{Survival prediction results}

\begin{table*}[ht]
\centering
\setlength{\tabcolsep}{4pt}
{\scriptsize
\begin{tabular}{ccccc}
 \toprule
 Architecture & IDC & CCRCC &  PRCC & LUAD \\
 \midrule
 ABMIL \cite{ABMIL} & 0.487 $\pm$ 0.079  &   0.561 $\pm$ 0.074 &  0.671 $\pm$ 0.076 & 0.584 $\pm$ 0.054 \\
DeepAttnMISL\cite{yao2020whole} & 0.472 $\pm$ 0.023 & 0.521 $\pm$ 0.084 &  0.472 $\pm$ 0.162 &  0.563 $\pm$ 0.037\\
GCN-MIL\cite{li2018graph,zhao2020predicting} & 0.534 $\pm$ 0.060 & 0.591 $\pm$ 0.093 & 0.636 $\pm$ 0.066 & 0.592 $\pm$ 0.070 \\
DS-MIL\cite{li2021dual} & 0.472 $\pm$ 0.020 & 0.548 $\pm$ 0.057  & 0.654 $\pm$ 0.134 & 0.537 $\pm$ 0.061\\
HIPT\cite{hipt} & 0.634 $\pm$ 0.050 & 0.642 $\pm$ 0.028 & 0.670 $\pm$ 0.065 & 0.538 $\pm$ 0.044 \\
HIPTLE & {\bf 0.636 $\pm$ 0.061} &   {\bf 0.684 $\pm$ 0.038} & {\bf 0.702 $\pm$ 0.083} & {\bf 0.611 $\pm$ 0.061} \\
\bottomrule
\end{tabular}
}
\caption{Experimental results comparing existing WSI classification algorithms against the  medium global structure, no level 2 pre-training configuration for survival prediction on TCGA datasets. The datasets used include the IDC subtype from TCGA-BRCA, CCRCC subtype from TCGA-KIRC, PRCC subtype from TCGA-KIRP, and LUAD subtype from TCGA-LUAD. The results for the first 5 algorithms above are taken from the HIPT paper\cite{hipt}. The best result in each column is highlighted using bold font.}
\label{Table:results_survival}
\end{table*}

In this experiment we compare our HIPTLE model against other existing models including the original HIPT approach~\cite{hipt}. We report the results for the following datasets: IDC cancer subtype from TCGA-BRCA; CCRCC cancer subtype from TCGA-KIRC; PRCC cancer subtype from TCGA-KIRP, and LUAD cancer subtype from TCGA-LUAD. We perform the experiments using 5 fold cross validation with the same splits used in the HIPT paper~\cite{hipt}. The results again show that HIPTLE outperforms all existing models including HIPT. It is encouraging to see the superior performance of HIPTLE carries over from WSI classification to survival prediction. 

\subsection{Varying pre-training and global structure with small training dataset}

\begin{table*}[ht]
\centering
\setlength{\tabcolsep}{4pt}
{\scriptsize
\begin{tabular}{cccc}
 \toprule
  & Most L2 PT & Med L2 PT &  No L2 PT \\
 \midrule
 Most global structure &  0.695 $\pm$ 0.086 & 0.812 $\pm$ 0.108  &  0.822 $\pm$ 0.110 \\
Med global structure & 0.819 $\pm$ 0.041 & 0.863 $\pm$ 0.073 & {\bf 0.864 $\pm$ 0.060}\\
No global structure (Max-MIL) \cite{CLAM} & - & -  & 0.828 $\pm$ 0.076 \\
\bottomrule
\end{tabular}
}
\caption{Results from varying the amount of level 2 pre-training and the amount of global structure when the training set is reduced to just 25\% of the training set size of the full TCGA-BRCA dataset. The best results in each column is highlighted using bold font.}
\label{Table:vary_pre_train_global_25_train}
\end{table*}

In this section we vary the amount of level 2 pre-training and the amount of global structure when the training dataset is reduced to just 25\% of the full TCGA-BRCA dataset. The classification task for this set of experiments is BRCA subtyping. The results once again show that the HIPTLE configuration (medium global structure and no level 2 pre-training) performs the best, and this continues to hold when the training dataset is small. This means that when there is limited training data available for the downstream task, heavier pre-training of the level 2 encoder still does not help.  

The medium global structure consistently performs better than most global structure and no global structure, which is consistent with earlier results (see Section \ref{sec:vary_pre-traing_global}).

\section{Conclusion}

The current trend for WSI classification is to propose complex methods that incorporate a more global view of the entire slide. In this paper we showed that instead focusing on the most predictive local patch can give results very similar to complex algorithms incorporating global structure. Rather than increasing the amount of global structure, we found that the key to high performance is using the appropriate level 1 encoder feature vectors. This suggests the models actually get most of their predictive power from the local patch level, which contains cell and local tissue micro-environment level information.

A very important finding of this paper is that the data used for pre-training the level 1 encoder matters a lot. Pre-training using a large dataset spanning 33 cancers actually works considerably worse than pre-training using a more focused set of 7 cancers (including the target cancer). This can be explained by the more efficient use of the representation space when only 7 cancers are used for pre-training. In fact, pre-training just on the target cancer gives very similar performance to using 7 cancers.

All the experiments show the HIPTLE model configuration consistently outperforms all other methods in all situations tested (including both WSI classification and survival prediction). HIPTLE uses a medium amount of global structure with no pre-training for the level 2 encoder and using level 1 encoder trained on 7 cancers. The robustness of these results shows that HIPTLE should be the first-choice model used in most WSI classification and survival prediction situations.

As future work we intend to explore predicting results of genetic tests like the BRCA gene for breast cancer and microsatellite instability (MSI) status for colorectal cancer (CRC). We would also like to perform a more in-depth study of survival prediction, involving more datasets using the various model configurations. Finally, given how important the level 1 encoder is to final performance, we would like to explore new novel methods for unsupervised pre-training of the level 1 encoder.

\section{Acknowledgements}
The results in this paper are in whole or part based upon data generated by the TCGA Research Network: https://www.cancer.gov/tcga.

\bibliographystyle{elsarticle-num} 
\bibliography{bibliography}

\begin{thebibliography}{10}
\expandafter\ifx\csname url\endcsname\relax
  \def\url#1{\texttt{#1}}\fi
\expandafter\ifx\csname urlprefix\endcsname\relax\def\urlprefix{URL }\fi
\expandafter\ifx\csname href\endcsname\relax
  \def\href#1#2{#2} \def\path#1{#1}\fi

\bibitem{berbis2023computational}
M.~A. Berb{\'\i}s, D.~S. McClintock, A.~Bychkov, J.~Van~der Laak,
  L.~Pantanowitz, J.~K. Lennerz, J.~Y. Cheng, B.~Delahunt, L.~Egevad, C.~Eloy,
  et~al., Computational pathology in 2030: a delphi study forecasting the role
  of ai in pathology within the next decade, EBioMedicine 88 (2023).

\bibitem{hipt}
R.~J. Chen, C.~Chen, Y.~Li, T.~Y. Chen, A.~D. Trister, R.~G. Krishnan,
  F.~Mahmood, Scaling vision transformers to gigapixel images via hierarchical
  self-supervised learning, in: Proceedings of CVPR, 2022, pp. 16144--16155.

\bibitem{adnan2020graph}
M.~Adnan, S.~Kalra, H.~R. Tizhoosh, Representation learning of histopathology
  images using graph neural networks, in: Proceedings of CVPR Workshops, 2020,
  pp. 988--989.

\bibitem{guan2022node}
Y.~Guan, J.~Zhang, K.~Tian, S.~Yang, P.~Dong, J.~Xiang, W.~Yang, J.~Huang,
  Y.~Zhang, X.~Han, Node-aligned graph convolutional network for whole-slide
  image representation and classification, in: Proceedings of CVPR, 2022, pp.
  18813--18823.

\bibitem{stegmuller2023scorenet}
T.~Stegm{\"u}ller, B.~Bozorgtabar, A.~Spahr, J.-P. Thiran, Scorenet: Learning
  non-uniform attention and augmentation for transformer-based
  histopathological image classification, in: Proceedings of the IEEE/CVF
  Winter Conference on Applications of Computer Vision, 2023, pp. 6170--6179.

\bibitem{li2021dual}
B.~Li, Y.~Li, K.~W. Eliceiri, Dual-stream multiple instance learning network
  for whole slide image classification with self-supervised contrastive
  learning, in: Proceedings of CVPR, 2021, pp. 14318--14328.

\bibitem{shao2021transmil}
Z.~Shao, H.~Bian, Y.~Chen, Y.~Wang, J.~Zhang, X.~Ji, et~al., Transmil:
  Transformer based correlated multiple instance learning for whole slide image
  classification, Advances in neural information processing systems 34 (2021)
  2136--2147.

\bibitem{vu2023handcrafted}
Q.~D. Vu, K.~Rajpoot, S.~E.~A. Raza, N.~Rajpoot, Handcrafted histological
  transformer (h2t): Unsupervised representation of whole slide images, Medical
  Image Analysis (2023) 102743.

\bibitem{CLAM}
M.~Y. Lu, D.~F. Williamson, T.~Y. Chen, R.~J. Chen, M.~Barbieri, F.~Mahmood,
  Data-efficient and weakly supervised computational pathology on whole-slide
  images, Nature biomedical engineering 5~(6) (2021) 555--570.

\bibitem{dino}
M.~Caron, H.~Touvron, I.~Misra, H.~J{\'e}gou, J.~Mairal, P.~Bojanowski,
  A.~Joulin, Emerging properties in self-supervised vision transformers, in:
  Proceedings of CVPR, 2021, pp. 9650--9660.

\bibitem{campanella2018terabyte}
G.~Campanella, V.~W.~K. Silva, T.~J. Fuchs, Terabyte-scale deep multiple
  instance learning for classification and localization in pathology, arXiv
  preprint arXiv:1805.06983 (2018).

\bibitem{coudray2018classification}
N.~Coudray, P.~S. Ocampo, T.~Sakellaropoulos, N.~Narula, M.~Snuderl,
  D.~Feny{\"o}, A.~L. Moreira, N.~Razavian, A.~Tsirigos, Classification and
  mutation prediction from non--small cell lung cancer histopathology images
  using deep learning, Nature medicine 24~(10) (2018) 1559--1567.

\bibitem{campanella2019clinical-NatureMed}
G.~Campanella, M.~G. Hanna, L.~Geneslaw, A.~Miraflor, V.~Werneck Krauss~Silva,
  K.~J. Busam, E.~Brogi, V.~E. Reuter, D.~S. Klimstra, T.~J. Fuchs,
  Clinical-grade computational pathology using weakly supervised deep learning
  on whole slide images, Nature medicine 25~(8) (2019) 1301--1309.

\bibitem{hou2016patch}
L.~Hou, D.~Samaras, T.~M. Kurc, Y.~Gao, J.~E. Davis, J.~H. Saltz, Patch-based
  convolutional neural network for whole slide tissue image classification, in:
  Proceedings of CVPR, 2016, pp. 2424--2433.

\bibitem{wang2016deep}
D.~Wang, A.~Khosla, R.~Gargeya, H.~Irshad, A.~H. Beck, Deep learning for
  identifying metastatic breast cancer, arXiv preprint arXiv:1606.05718 (2016).

\bibitem{ABMIL}
M.~Ilse, J.~Tomczak, M.~Welling, Attention-based deep multiple instance
  learning, in: International conference on machine learning, PMLR, 2018, pp.
  2127--2136.

\bibitem{zhang2022dtfd}
H.~Zhang, Y.~Meng, Y.~Zhao, Y.~Qiao, X.~Yang, S.~E. Coupland, Y.~Zheng,
  Dtfd-mil: Double-tier feature distillation multiple instance learning for
  histopathology whole slide image classification, in: Proceedings of CVPR,
  2022, pp. 18802--18812.

\bibitem{VIT}
A.~Dosovitskiy, L.~Beyer, A.~Kolesnikov, D.~Weissenborn, X.~Zhai,
  T.~Unterthiner, M.~Dehghani, M.~Minderer, G.~Heigold, S.~Gelly, et~al., An
  image is worth 16x16 words: Transformers for image recognition at scale, ICLR
  (2021).

\bibitem{bejnordi2017diagnostic}
B.~E. Bejnordi, M.~Veta, P.~J. Van~Diest, B.~Van~Ginneken, N.~Karssemeijer,
  G.~Litjens, J.~A. Van Der~Laak, M.~Hermsen, Q.~F. Manson, M.~Balkenhol,
  et~al., Diagnostic assessment of deep learning algorithms for detection of
  lymph node metastases in women with breast cancer, Jama 318~(22) (2017)
  2199--2210.

\bibitem{kim2021paip}
Y.~J. Kim, H.~Jang, K.~Lee, S.~Park, S.-G. Min, C.~Hong, J.~H. Park, K.~Lee,
  J.~Kim, W.~Hong, et~al., Paip 2019: Liver cancer segmentation challenge,
  Medical image analysis 67 (2021) 101854.

\bibitem{macenko}
M.~Macenko, M.~Niethammer, J.~S. Marron, D.~Borland, J.~T. Woosley, X.~Guan,
  C.~Schmitt, N.~E. Thomas, A method for normalizing histology slides for
  quantitative analysis, in: 2009 IEEE international symposium on biomedical
  imaging: from nano to macro, IEEE, 2009, pp. 1107--1110.

\bibitem{adamw}
I.~Loshchilov, F.~Hutter, Decoupled weight decay regularization, arXiv preprint
  arXiv:1711.05101 (2017).

\bibitem{kingma2014adam}
D.~P. Kingma, J.~Ba, Adam: A method for stochastic optimization, arXiv preprint
  arXiv:1412.6980 (2014).

\bibitem{yao2020whole}
J.~Yao, X.~Zhu, J.~Jonnagaddala, N.~Hawkins, J.~Huang, Whole slide images based
  cancer survival prediction using attention guided deep multiple instance
  learning networks, Medical Image Analysis 65 (2020) 101789.

\bibitem{li2018graph}
R.~Li, J.~Yao, X.~Zhu, Y.~Li, J.~Huang, Graph cnn for survival analysis on
  whole slide pathological images, in: International Conference on Medical
  Image Computing and Computer-Assisted Intervention, Springer, 2018, pp.
  174--182.

\bibitem{zhao2020predicting}
Y.~Zhao, F.~Yang, Y.~Fang, H.~Liu, N.~Zhou, J.~Zhang, J.~Sun, S.~Yang,
  B.~Menze, X.~Fan, et~al., Predicting lymph node metastasis using
  histopathological images based on multiple instance learning with deep graph
  convolution, in: Proceedings of CVPR, 2020, pp. 4837--4846.

\end{thebibliography}

\end{document}